\documentclass[]{fairmeta}
\title{Small Vision-Language Models are Smart Compressors for Long Video Understanding}

\author[1,2,*,\dagger]{Junjie Fei}
\author[1]{Jun Chen}
\author[1]{Zechun Liu}
\author[1]{Yunyang Xiong}
\author[1]{Chong Zhou}
\author[1]{Wei Wen}
\author[1]{Junlin Han}
\author[1,2]{\\Mingchen Zhuge}
\author[1]{Saksham Suri}
\author[1]{Qi Qian}
\author[1,2]{Shuming Liu}
\author[1]{Lemeng Wu}
\author[1]{Raghuraman Krishnamoorthi}
\author[1,\dagger]{Vikas Chandra}
\author[2,\dagger]{Mohamed Elhoseiny}
\author[1,\dagger]{Chenchen Zhu}

\affiliation[1]{Meta AI}
\affiliation[2]{King Abdullah University of Science and Technology (KAUST)}

\contribution[*]{Work done at Meta}
\contribution[\dagger]{Project lead}

\abstract{Adapting Multimodal Large Language Models (MLLMs) for hour-long video understanding is severely bottlenecked by context window limits.
Dense visual streams quickly saturate input token budgets and exacerbate the \emph{lost-in-the-middle} phenomenon. Existing efficiency heuristics, such as sparse sampling or query-agnostic uniform pooling, blindly sacrifice fidelity. They frequently discard transient decisive moments, blur fine-grained evidence, and waste representational bandwidth on irrelevant backgrounds.
In this paper, we propose \textbf{Tempo}, an efficient, query-aware framework that compresses long videos for downstream understanding. Tempo leverages a Small Vision-Language Model (SVLM) to act as a local temporal compressor. It casts visual token reduction as an early cross-modal distillation process, generating compact, intent-aligned video representations in a single forward pass.
To enforce strict inference budgets without breaking causality, we introduce \textbf{Adaptive Token Allocation (ATA)}. Exploiting the SVLM's inherent zero-shot relevance prior and empirical semantic front-loading, ATA acts as a training-free, $O(1)$ dynamic router. It allocates dense bandwidth to query-critical segments while compressing redundancies into minimal \emph{temporal anchors} to maintain the global storyline.
Extensive experiments demonstrate that our compact 6B architecture achieves state-of-the-art performance with aggressive dynamic compression (0.5--16 tokens/frame). On the extreme-long LVBench (4101s), Tempo scores \textbf{52.3} under a strict 8K visual token budget, outperforming proprietary baselines such as GPT-4o and Gemini 1.5 Pro. Scaling to 2048 frames pushes performance to \textbf{53.7}.
Crucially, empirical profiling reveals that Tempo frequently compresses hour-long videos to token counts substantially below theoretical computational limits, proving that true long-form video understanding relies on intent-driven efficiency rather than greedily padded context windows.}

\date{\today}
\correspondence{\email{junjiefei@outlook.com}, \email{chenchenz@meta.com}}

\metadata[Code]{\url{https://github.com/FeiElysia/Tempo}}
\metadata[Project \& Demo]{\url{https://FeiElysia.github.io/tempo-page/}}

\usepackage{xspace}
\newcommand{\ie}{\emph{i.e.}\xspace}
\newcommand{\eg}{\emph{e.g.}\xspace}

\usepackage{booktabs}
\usepackage{multirow}
\usepackage[table]{xcolor}
\usepackage{array}
\usepackage{amssymb}
\usepackage{graphicx}
\usepackage[ruled,vlined]{algorithm2e}
\usepackage{pifont}
\usepackage{fontawesome5}

\begin{document}

\maketitle

% ---------------------------------------------------------------
\begin{figure}[t]
    \centering
    \includegraphics[width=\textwidth]{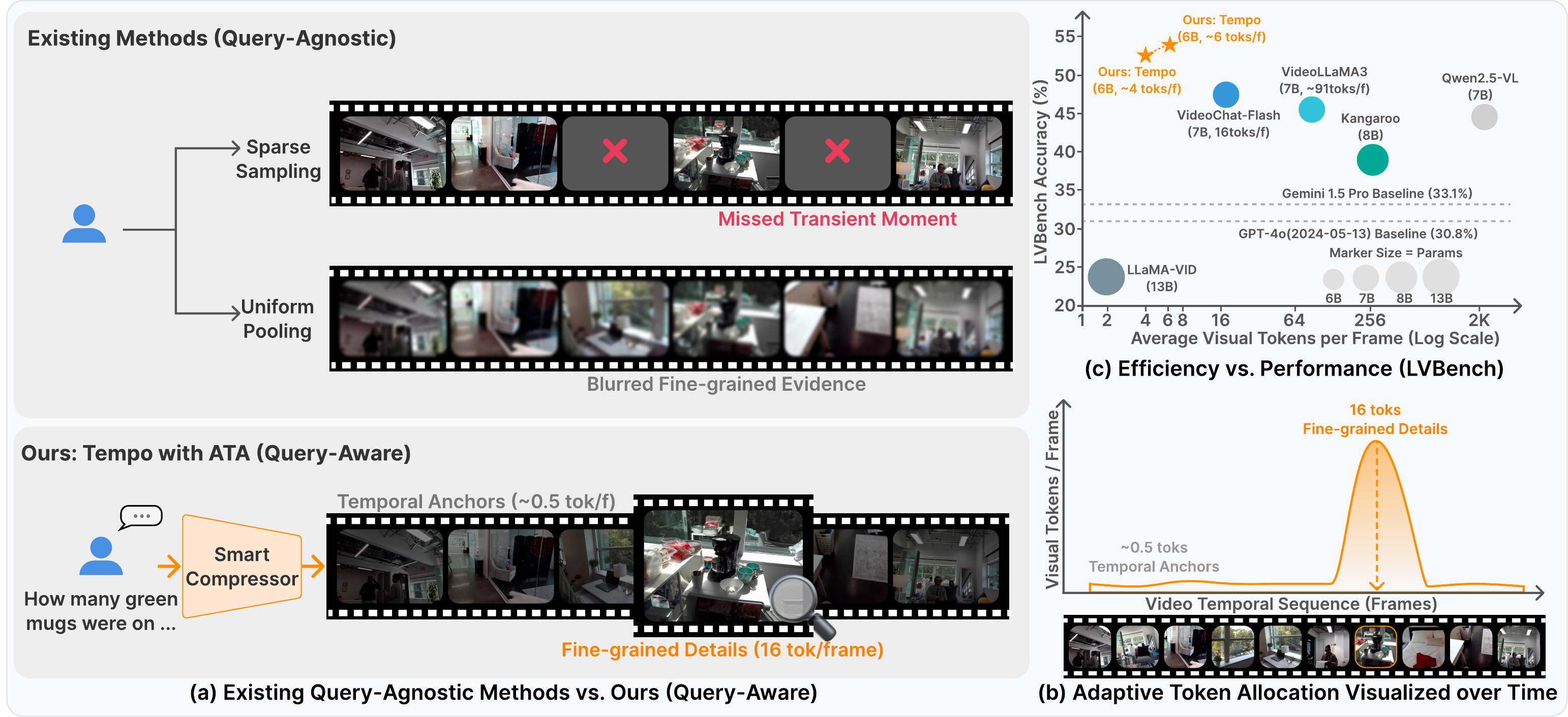}
    \caption{
    \textbf{Tempo achieves SOTA long video understanding via query-aware Adaptive Token Allocation (ATA).}
    \textbf{(a) Motivation:} Query-agnostic methods either miss transient moments (sparse sampling) or blur details (uniform pooling). Tempo instead utilizes a small vision-language model as a \emph{smart compressor} for query-aware cross-modal distillation.
    \textbf{(b) Mechanism:} ATA dynamically allocates high bandwidth (16 tokens/frame) to relevant segment for fine-grained details, while compressing redundant contexts into minimal temporal anchors ($\sim$0.5 tokens/frame) to maintain causality.
    \textbf{(c) Result:} Leading performance on LVBench. Tempo-6B achieves superior accuracy at extreme compression rates (\ie, 4 or 6 tokens/frame), outperforming open-source models and proprietary baselines with a fraction of the context budget.
    }
    % \vspace{-1em}
    \label{fig:teaser}
\end{figure}
\section{Introduction}
\label{sec:intro}

The advancement of Multimodal Large Language Models (MLLMs) has significantly transformed visual understanding, empowering systems to perform complex semantic analysis over images and short video clips~\cite{liu2023visual,liu2024improved,zhu2023minigpt,li2024llava,bai2025qwen3,li2025videochat,zhang2024llava,zhang2023video}. However, scaling these capabilities to hour-long videos remains challenging.
The core difficulty lies in the structural mismatch between the massive, continuous visual stream of long videos and the rigidly bounded context windows of downstream LLMs.
As temporal duration expands, raw visual tokens quickly overwhelm the input capacity, severely diluting attention mechanisms and causing models to fail at retrieving sparse evidence buried within extensive contexts~\cite{liu2024lost}.

To fit long video understanding into limited contexts, existing methods typically force one of two compromises.
A common approach is sparse frame sampling~\cite{xu2024pllava,li2025videochat,lin2024video}, which reduces compute but inevitably risks skipping the transient yet decisive moments required to answer a specific query.
Alternatively, methods retain more frames but apply query-agnostic compression, such as uniform spatiotemporal pooling~\cite{maaz2024video,jiang2025storm} or token merging~\cite{bolya2022token,li2024videochat,jin2024chat}.
By compressing without knowing \emph{what the user will ask}, these heuristics often blur fine-grained evidence in query-critical segments while wasting representational bandwidth on irrelevant backgrounds.
In essence, most existing pipelines reduce visual evidence before interacting with the language model, preventing the dynamic allocation of bandwidth to query-critical segments.
Even pioneering query-aware approaches (\eg, LongVU~\cite{shen2024longvu}) rely on disjoint auxiliary feature-matching modules, thereby decoupling the routing mechanism from the end-to-end multimodal pipeline.

We introduce \textbf{Tempo}, an efficient query-aware framework for long video understanding that natively learns to compress videos for downstream text generation tasks.
As its name suggests, Tempo acts as an intelligent temporal compressor that dynamically distributes the \emph{rhythm} of the video: it allocates high token bandwidth to semantic beats relevant to the query while swiftly fast-forwarding through redundant contexts.
Rather than treating visual compression as a purely visual, query-agnostic operation~\cite{jiang2025storm,li2024videochat}, Tempo casts this reduction as an early cross-modal semantic distillation process.
Concretely, Tempo leverages a Small Vision-Language Model (SVLM) as a local compressor, seamlessly bridging it with an LLM for global understanding and response generation.
By prepending the user query to the SVLM input, Tempo performs a preliminary cross-modal distillation pass that produces compact video memory tokens aligned with the user's intent and is trained end-to-end with standard auto-regressive objectives.

A practical challenge is enforcing a strict token budget at inference time (\eg, representing a 1024-frame video under an 8K visual token budget) without sacrificing either fine-grained evidence or global causal structure.
To this end, we propose \textbf{Adaptive Token Allocation (ATA)}, a training-free inference strategy guided by two key empirical properties of the Tempo architecture.
(i) \emph{Zero-shot relevance prior and temporal anchors.}
Inheriting from the base model's extensive multimodal pre-training, the local compressor exhibits a zero-shot ability to estimate query-video relevance without auxiliary supervision.
ATA exploits this prior to allocate budgets segment-wise, enabling an aggressive dynamic compression range (0.5--16 tokens per frame). Crucially, instead of hard pruning, which breaks causality, ATA preserves dense representational bandwidth for relevant segments while compressing redundant contexts into minimal temporal anchors (\ie, 4 tokens) to maintain the global storyline.
(ii) \emph{Semantic front-loading driven by causal attention.}
Our ablations empirically reveal that under the SVLM's causal attention, salient visual semantics natively concentrate into the earliest video memory tokens.
Consequently, a simple head truncation effectively isolates high-value evidence, avoiding lossy spatial blurring with zero overhead.

In summary, our contributions are:
\vspace{-0.5em}
\begin{itemize}
    \item \textbf{Tempo:} an end-to-end, query-aware compression framework for long video understanding. It directly addresses the context window bottleneck by unifying an SVLM-based local compressor and an LLM-based global decoder, performing query-conditioned cross-modal distillation in a single forward pass.
    \item \textbf{ATA:} a training-free, budget-aware inference strategy leveraging the local compressor's inherent zero-shot relevance prior and semantic front-loading. ATA dynamically dictates the optimal token allocation, preserving fine-grained details for query-critical moments while compressing redundancies into minimal temporal anchors to maintain global causal structure.   
    \item \textbf{Scaling Behaviors:} an empirical analysis revealing that optimal resource allocation varies with the task and video duration. While a 4K visual token budget acts as a \emph{sweet spot} for standard long video tasks (\eg, Video-MME Long, \textbf{30--60 mins}), restrictive budgets ultimately limit performance on extreme-long videos (\eg, LVBench, \textbf{$>$1 hour}). Scaling to larger capacities unlocks new performance peaks. Notably, in practice we observe that Tempo allocates tokens largely based on semantic necessity, often compressing hour-long videos far below the available token budget.
    \item \textbf{Leading Performances:} despite being a compact 6B model, Tempo sets a new state-of-the-art across long video benchmarks. On challenging LVBench, it scores \textbf{52.3} under a 8K budget, outperforming proprietary baseline (\eg, GPT-4o, Gemini 1.5 Pro) and open-source counterparts (\eg, VideoChat-Flash). Scaling to 2048 frames with a 12K budget further pushes performance to \textbf{53.7}, demonstrating robust hour-long video understanding of our proposed Tempo.
\end{itemize}
\section{Related Work}
\label{sec:related_work}

\subsection{Multimodal Large Language Models for Videos}
\label{sec:related_work:MLLMs}

The rapid evolution of MLLMs has established a dominant paradigm: aligning pre-trained visual encoders with powerful LLMs. Recent state-of-the-art models, such as VideoChat2~\cite{li2025videochat}, VILA~\cite{lin2024vila}, LLaVA-OneVision~\cite{li2024llava}, VITA-1.5~\cite{fu2025vita}, Kimi-VL~\cite{team2025kimi}, InternVL3.5~\cite{wang2025internvl3}, Molmo2~\cite{clark2026molmo2}, and the Qwen-VL series~\cite{bai2025qwen3},
demonstrate exceptional capability in short video understanding. They typically map sampled video frames directly into the LLM's context window. While effective for short-horizon tasks, extending this dense representation to hour-long videos results in a linear explosion of visual tokens. This quickly overwhelms the maximum context length of the LLM, leading to prohibitive computational costs and exacerbating the \emph{lost-in-the-middle} phenomenon~\cite{liu2024lost}, where models fail to retrieve pivotal evidence buried in extensive multimodal contexts.

\subsection{Context Extension and Token Reduction}
\label{sec:related_work:reduction}

To comprehend extended temporal horizons, recent efforts generally bifurcate into two directions. The first direction focuses on context extension via algorithmic extrapolation, architectural innovations, or system-level parallelization to natively support massive token sequences.
For instance, LongVA~\cite{zhang2024long} extrapolates the context window to comprehend extensive visual tokens, LongVILA~\cite{chen2024longvila} introduces sequence parallelism for long-context training, and LongLLaVA~\cite{wang2024longllava} employs a hybrid Mamba-Transformer architecture to mitigate memory constraints.
While these approaches successfully preserve visual fidelity and push the context boundaries, they strictly rely on processing dense visual streams. Consequently, ingesting hundreds of thousands of visual tokens per video still incurs exorbitant memory footprints and computational overhead, rendering them highly resource-intensive for routine inference.
The second, more prevalent direction relies on query-agnostic token reduction. Drawing inspiration from image-level token pruning and merging techniques like FastV~\cite{chen2024image} and ToMe~\cite{bolya2022token}, video MLLMs typically employ spatiotemporal pooling or fixed-rate sparse sampling~\cite{maaz2024video,li2025videochat,jin2024chat,li2024videochat,jiang2025storm}.
For instance, VideoChat-Flash~\cite{li2024videochat} leverages visual redundancy to hierarchically compress tokens, while Storm~\cite{jiang2025storm} applies temporal and spatial pooling to fit tight token budgets.
However, because these heuristics are completely agnostic to the user's textual query, they risk blurring semantic boundaries and discarding transient, fine-grained segments that may be critical to the downstream question.

\subsection{Hierarchical and Query-Aware Video Architectures}
\label{sec:related_work:query_aware}
To overcome uniform processing limits, models like SLOWFAST-LLAVA, LLaVA-Video$_{slowfast}$, and Keye-VL-1.5 deploy dual pathways to balance spatial and temporal resolutions~\cite{feichtenhofer2019slowfast,xu2024slowfast,zhang2024llava,yang2025kwai}. However, whether utilizing static sampling or dynamic inter-frame similarity, their resource allocation remains purely vision-driven and fundamentally detached from the user's textual intent.
Recent works have begun to explore query-aware processing~\cite{li2024llama,islam2025bimba,shen2024longvu}. BIMBA~\cite{islam2025bimba} introduces an optional query-conditioned token selection mechanism. LongVU~\cite{shen2024longvu} leverages cross-modal attention for selective spatial compression, yet still depends on disjoint auxiliary modules that decouple the routing mechanism from the end-to-end multimodal decoding process.
Tempo fundamentally advances this trajectory by natively employing an SVLM as an active, query-conditioned temporal compressor in a single forward pass. Furthermore, our ATA mechanism exploits the SVLM's inherent zero-shot relevance prior to dynamically dictate the video's \emph{rhythm}. This preserves dense, high-fidelity tokens for critical segments while compressing irrelevant backgrounds into minimal temporal anchors, achieving causal-preserving sequence assembly with zero routing overhead.

\section{TEMPO}
\label{sec:method}

\subsection{Overall}
\label{sec:method:overview}
\begin{figure}[t]
    \centering
    \includegraphics[width=\textwidth]{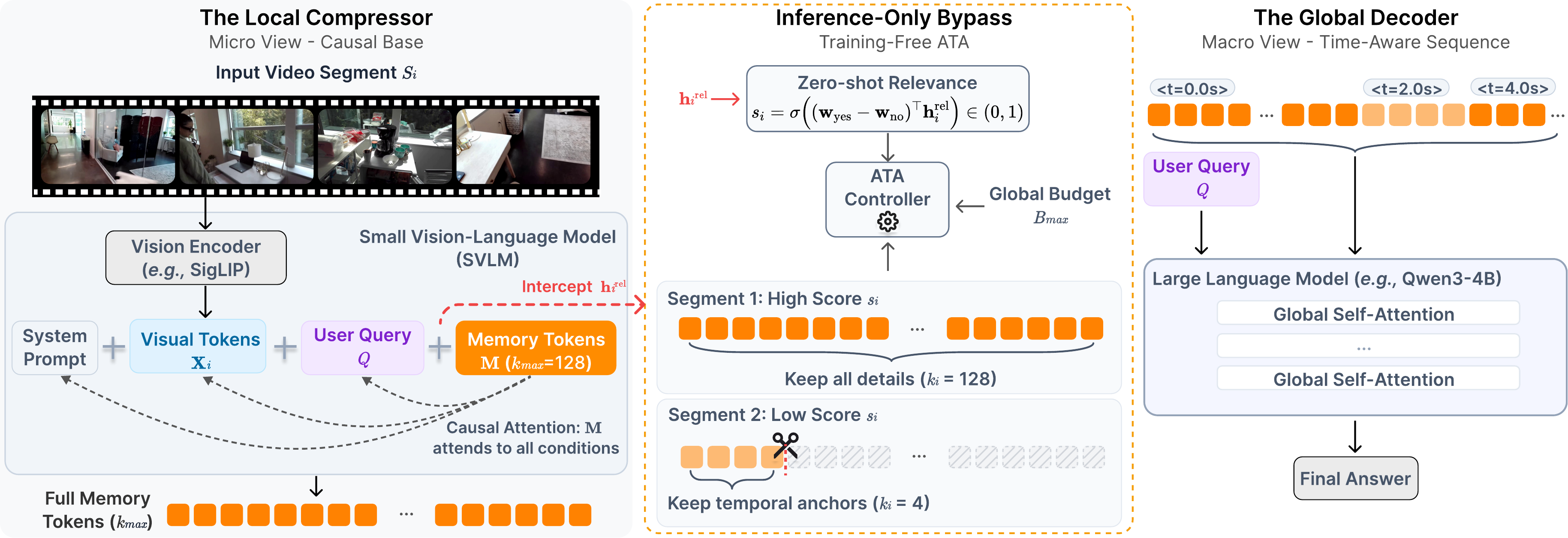}
    \caption{
        \textbf{Overview of the Tempo framework.} 
        Our unified architecture casts long video understanding as an end-to-end, query-aware compression process. 
        \textbf{The Local Compressor (Left).} For each segment, a Small Vision-Language Model (SVLM) acts as a semantic temporal compressor. Under causal attention, learnable memory tokens $\mathbf{M}$ inherently distill the preceding visual tokens $\mathbf{X}_i$ and user query $Q$. 
        \textbf{Inference-Only Bypass (Middle).} During a single forward pass, an Adaptive Token Allocation (ATA) controller intercepts the hidden state $\mathbf{h}_i^{\mathrm{rel}}$ to compute a zero-shot relevance score $s_i$. This enables an $\mathcal{O}(1)$ dynamic head truncation, allocating dense bandwidth to query-critical segments while compressing redundancies into minimal temporal anchors to strictly satisfy a global budget $B_{\max}$. 
        \textbf{The Global Decoder (Right).} The compressed memory tokens are assembled into a highly sparse, time-aware sequence using explicit temporal tags (\eg, \texttt{<t=2.0s>}). A global LLM synthesizes this condensed multimodal context to generate the final response.
    }
    % \vspace{-0.8em}
    \label{fig:tempo}
\end{figure}

We target the fundamental bottleneck in long video MLLMs: the downstream LLM can only attend to a limited number of visual tokens, while hour-long videos produce a massive, continuous stream. Tempo resolves this mismatch by turning visual token reduction into an early cross-modal distillation problem.

\paragraph{Problem Setup.}
Given a long video $\mathcal{V}$ and a user query $Q$, we uniformly partition $\mathcal{V}$ into $N$ temporal segments $\mathcal{S}=\{S_1,\dots,S_N\}$. Our goal is to convert each $S_i$ into a compact set of query-conditioned video memory tokens, with the total sequence bounded by a global inference budget~$B_{\max}$, enabling the downstream LLM to process the entire video and generate the final answer efficiently.

\paragraph{Architecture.}
Tempo constitutes a two-level generative hierarchy (Fig.~\ref{fig:tempo}):
(1) an SVLM-based local compressor $\mathcal{C}_\phi$, and (2) an LLM-based global decoder $\mathcal{D}_\theta$.
Concretely, the SVLM's native vision encoder maps segment $S_i$ to dense visual tokens $\mathbf{X}_i$. Its causal attention then performs query-conditioned distillation, integrating $\mathbf{X}_i$ and query $Q$ into learnable memory tokens $\mathbf{M}$.
This yields a fixed-capacity representation $\mathbf{H}_i$ of exactly $k_{\max}$ tokens. A linear projector maps $\mathbf{H}_i$ into the LLM's embedding space as $\tilde{\mathbf{H}}_i$. Finally, the global LLM $\mathcal{D}_\theta$ consumes all memory tokens $\{\tilde{\mathbf{H}}_i\}_{i=1}^N$ alongside $Q$ to auto-regressively decode the answer.

\paragraph{Training vs. Inference.}
Tempo is trained with a fixed per-segment capacity $k_{\max}$ to learn a strong query-aware local compressor $\mathcal{C}_\phi$. At inference, we additionally enforce a global budget $B_{\max}$. We therefore introduce ATA, a training-free strategy that uses a zero-shot relevance prior extracted from the same SVLM forward pass to allocate per-segment budgets $k_i \in [k_{\min}, k_{\max}]$, followed by a constant-time head truncation.

\subsection{Query-Aware Visual Compression}
\label{sec:method:compression}

We cast segment compression as a query-driven sequence-to-sequence transformation. An explicit information bottleneck forces the $\mathcal{C}_\phi$ to discard visual redundancies and distill semantic evidence relevant to user intent.

\paragraph{SVLM Input Construction.}
For each segment $S_i$, the SVLM constructs a single causal sequence comprising: (i) a system prompt, (ii) visual tokens $\mathbf{X}_i$ (extracted via its native vision encoder), (iii) user query $Q$, and (iv) learnable memory tokens $\mathbf{M}$.
Placing $\mathbf{M}$ last is critical: under causal attention, each memory token inherently attends to all preceding visual and textual contexts. This conditions the SVLM to distill query-aligned evidence into $\mathbf{M}$. Extracting their final-layer hidden states yields the compressed representation $\mathbf{H}_i \in \mathbb{R}^{k_{\max}\times d_s}$.

\paragraph{Sequence Assembly \& Temporal Grounding.}
To preserve temporal identity and causal order across the entire video, we prepend an explicit textual timestamp (\eg, \texttt{<t=2.0s>}) to each segment when assembling the global context. In practice, these temporal tags significantly stabilize long-range attribution (\emph{\ie, which evidence comes from where}) within the downstream global LLM.

\paragraph{End-to-End Learning.}
Let the ground-truth answer be $A=\{a_t\}_{t=1}^{T}$. The global $\mathcal{D}_\theta$ receives all projected segment memories $\{\tilde{\mathbf{H}}_i\}_{i=1}^{N}$ in temporal order, optimized via standard auto-regressive next-token prediction:
\begin{equation}
    \mathcal{L}_{\mathrm{AR}}(\theta,\phi) = -\sum_{t=1}^{T}\log p_{\theta}\big(a_t \mid a_{<t}, Q, \{\tilde{\mathbf{H}}_i\}_{i=1}^{N}\big)
    \label{eq:loss}
\end{equation}
Crucially, we do not impose auxiliary compression losses, routing networks, or heuristic token-dropping regularizations during training. The fixed capacity of $k_{\max}$ memory tokens acts as a hard structural bottleneck. The gradients back-propagated from $\mathcal{L}_{\mathrm{AR}}$ naturally compel the compressor $\mathcal{C}_\phi$ to discard query-irrelevant backgrounds and pack the most predictive visual evidence into this bounded space.

\subsection{Zero-Shot Relevance Prior}
\label{sec:method:relevance}

A core insight driving Tempo is that modern multimodal foundation models (\eg, Qwen3-VL~\cite{bai2025qwen3}) inherently possess a robust zero-shot capability to evaluate semantic alignment between a visual sequence and a text query (Refer to Sec.~\ref{sec:exp:ablation} -- D). We harness this foundational prior to extract a highly accurate relevance signal without introducing or training any auxiliary routing modules.

\paragraph{Logit-Based Relevance Score.}
To explicitly elicit this prior during inference, we slightly augment our training system prompt. Following the standard compression instruction, we append a strict binary directive: \emph{``Now, before compressing, answer exactly `Yes' or `No': is this segment relevant to the Query?''} Let $\mathbf{h}_i^{\mathrm{rel}} \in \mathbb{R}^{d_s}$ be the final hidden state immediately preceding the model's binary response. Using the SVLM's frozen language modeling head weights for the vocabulary tokens \texttt{Yes} ($\mathbf{w}_{\mathrm{yes}}$) and \texttt{No} ($\mathbf{w}_{\mathrm{no}}$), we compute a continuous relevance probability $s_i$ via logit difference~\cite{li2026qwen3}:
\begin{equation}
s_i = \sigma\Big( (\mathbf{w}_{\mathrm{yes}}-\mathbf{w}_{\mathrm{no}})^\top \mathbf{h}_i^{\mathrm{rel}} \Big) \in (0,1),
\label{eq:relevance}
\end{equation}
where $\sigma(\cdot)$ is the \texttt{Sigmoid} function. This $O(1)$ projection avoids auto-regressive decoding overhead while yielding a highly stable ranking signal.

\paragraph{Single-Pass Design.}
The score $s_i$ and the compressed memory tokens $\mathbf{H}_i$ are extracted within a single forward pass of $\mathcal{C}_\phi$. As illustrated in the \emph{Inference-Only Bypass} (Fig.~\ref{fig:tempo}), we simply intercept the hidden state $\mathbf{h}_i^{\mathrm{rel}}$ to compute the zero-shot score, and then seamlessly continue the forward pass to extract $\mathbf{H}_i$. This architectural elegance guarantees that both the relevance routing signal and the compressed representations are rigorously conditioned on the exact same multimodal context, achieving adaptive evaluation with effectively zero latency.

\subsection{Adaptive Token Allocation (ATA)}
\label{sec:method:ata}

At inference, the total visual context provided to the global LLM must strictly satisfy a bounded capacity~$B_{\max}$. As illustrated in Fig.~\ref{fig:tempo}, the ATA controller translates the zero-shot scores $\{s_i\}$ into dynamic per-segment token budgets $k_i$, executing the physical compression via zero-overhead head truncation.

\paragraph{Stage 1: Contrastive Linear Allocation.}
To guarantee causal continuity across the entire video sequence, we enforce a minimal temporal anchor for every segment, regardless of its relevance. We first normalize the raw scores via Min-Max scaling: $\hat{s}_i = (s_i-\min(\mathbf{s})) / (\max(\mathbf{s})-\min(\mathbf{s})+\epsilon)$. To maximize the contrast between query-critical events and irrelevant backgrounds, we linearly map these normalized scores to a target capacity:
\begin{equation}
k_i^{\mathrm{ideal}} = k_{\min} + \lfloor (k_{\max}-k_{\min}) \cdot \hat{s}_i \rfloor.
\label{eq:ideal_alloc}
\end{equation}

\paragraph{Stage 2: Capacity-Aware Protection.}
Let $B_{\mathrm{base}}=N\cdot k_{\min}$ represent the foundational cost required to maintain the global temporal anchors. If the sum of ideal allocations satisfies the global limit ($\sum_i k_i^{\mathrm{ideal}} \le B_{\max}$), we directly adopt $\{k_i^{\mathrm{ideal}}\}$ to maximize sparsity. Otherwise, we distribute the residual budget $B_{\mathrm{res}}=B_{\max}-B_{\mathrm{base}}$ proportionally based on the normalized scores:
\begin{equation}
k_i = k_{\min} + \left\lfloor B_{\mathrm{res}} \cdot \frac{\hat{s}_i}{\sum_{j=1}^{N}\hat{s}_j+\epsilon} \right\rfloor.
\label{eq:soft_alloc}
\end{equation}
We then discretize $\{k_i\}$ and distribute any fractional remainders to strictly ensure $\sum_i k_i \le B_{\max}$ (Alg.~\ref{alg:ata}).

\begin{algorithm}[t]
\caption{Adaptive Token Allocation (ATA) at inference}
\label{alg:ata}
\KwIn{Segment memories $\{\mathbf{H}_i\}_{i=1}^N$, relevance scores $\{s_i\}_{i=1}^N$, budget $B_{\max}$, bounds $k_{\min},k_{\max}$}
\KwOut{Budgeted memories $\{\mathbf{H}_i^{\mathrm{ATA}}\}_{i=1}^N$}
Normalize $\{s_i\}\rightarrow\{\hat{s}_i\}$ by min-max scaling\;
Compute $k_i^{\mathrm{ideal}}$ via Eq.~\eqref{eq:ideal_alloc}\;
\eIf{$\sum_i k_i^{\mathrm{ideal}} \le B_{\max}$}{
$k_i \leftarrow k_i^{\mathrm{ideal}}$\;
}{
$k_i \leftarrow$ Eq.~\eqref{eq:soft_alloc} with $B_{\mathrm{base}}=N k_{\min}$\;
}
Discretize $\{k_i\}$ to integers s.t.\ $\sum_i k_i \le B_{\max}$\;
$\mathbf{H}_i^{\mathrm{ATA}} \leftarrow \mathbf{H}_i[1{:}k_i], \quad \forall i \in \{1,\dots,N\}$\;
\Return{$\{\mathbf{H}_i^{\mathrm{ATA}}\}_{i=1}^N$}\;
\end{algorithm}

\paragraph{Head Truncation: Zero-Overhead Token Selection.}
Once the dynamic budget $k_i$ is allocated, we compress the segment by simply slicing the memory sequence, \ie, $\mathbf{H}_i^{\mathrm{ATA}} = \mathbf{H}_i[1{:}k_i]$. Driven by the auto-regressive nature of the SVLM's causal attention, we empirically observe a semantic front-loading phenomenon: the local compressor packs the most salient global evidence into the earliest generated memory tokens (Refer to Sec.~\ref{sec:exp:ablation} -- C). Consequently, this $O(1)$ tensor slice naturally isolates high-value semantics without introducing lossy spatiotemporal pooling. The final global sequence $\{\tilde{\mathbf{H}}_i^{\mathrm{ATA}}\}_{i=1}^{N}$ strictly conforms to $B_{\max}$, rendering memory footprints entirely predictable even for hour-long reasoning. 
\section{Experiments}
\label{sec:experiments}

\subsection{Experimental Setup}
\label{sec:exp:setup}

\paragraph{Architecture \& Implementation.}
Tempo's local SVLM is initialized from Qwen3-VL-2B-Instruct, while the global LLM uses Qwen3-LM-4B. A linear projector bridges the SVLM's memory space to the LLM, yielding a compact 6B-parameter architecture.
We extract frames at 2 FPS via Decord, applying uniform subsampling if limits are exceeded. During training, continuous videos are partitioned into 4-frame segments, each compressed by the SVLM into $k_{\max}=128$ memory tokens. During inference, we expand the segment window to 8 frames. ATA (Sec.~\ref{sec:method:ata}) strictly enforces a global visual budget $B_{\max}$ (4K or 8K) via head truncation. Models are trained on a 64-GPU cluster with FSDP. Additional hyper-parameters are provided in the Appendix~\ref{supp:training_details}.

\paragraph{Progressive Training Curriculum.}
We adopt a rigorous four-stage progressive training curriculum to ensure stable optimization and context extrapolation:
\begin{itemize}
    \item \textbf{Stage 0 (Modality Alignment):} We freeze both the SVLM and the LLM, exclusively optimizing the linear projector on the standard LCS-558K dataset~\cite{liu2023visual}. This establishes the fundamental vision-language alignment, bridging the SVLM's visual representations with the LLM's text embedding.
    
    \item \textbf{Stage 1 (Pre-training):} We unfreeze the entire architecture and optimize it on a large-scale, curated multimodal corpus comprising $\sim$2M images, $\sim$1.38M videos, and $\sim$143K pure text samples. During this phase, videos are sparsely sampled at 8 frames, endowing the model with initial temporal perception.
    
    \item \textbf{Stage 2 (Broad Supervised Fine-Tuning):} To develop robust instruction-following and semantic-aware temporal reasoning capabilities, we perform comprehensive SFT using a highly diverse data mixture ($\sim$0.93M images, $\sim$2.25M videos, and $\sim$71K text samples). In this stage, the temporal context is systematically expanded, with the maximum number of sampled frames per video strictly capped at 128.
    
    \item \textbf{Stage 3 (Long-Context SFT):} To effectively extrapolate the context window, we freeze the SVLM and exclusively fine-tune the global LLM on a high-quality subset of $\sim$384K samples from Stage 2. Here, the maximum frame limit is extended to 384, enabling the LLM to handle long temporal sequences.
    
\end{itemize}
To curate our training data, we primarily follow the data mixtures established by VideoChat-Flash~\cite{li2024videochat} and LLaVA-OneVision-1.5~\cite{an2025llava}. All training datasets utilized throughout our progressive curriculum are publicly accessible, ensuring full reproducibility.

\paragraph{Evaluation Benchmarks \& Baselines.}
To evaluate Tempo's long video understanding, we conduct comprehensive experiments across four prominent benchmarks, \ie, LongVideoBench~\cite{wu2024longvideobench}, MLVU~\cite{zhou2025mlvu}, Video-MME~\cite{fu2025video}, LVBench (extreme-long video)~\cite{wang2025lvbench}, spanning standard long-form tasks to hour-long stress tests. We benchmark Tempo against widely adopted proprietary baselines (\eg, GPT-4o, Gemini Pro 1.5), general open-weight MLLMs (\eg, InternVL, Qwen-VL), and specialized long-video MLLMs (\eg, VideoChat-Flash, LongVA). All evaluations are conducted using the \texttt{lmms-eval}.

\begin{table*}[t]
    \centering
    \setlength{\tabcolsep}{8pt} 
    \renewcommand{\arraystretch}{1.2} 
    \caption{\textbf{Comparison with state-of-the-art MLLMs on long video benchmarks, highlighting Tempo's superior accuracy and extreme token efficiency.} \textbf{Bold} and \underline{underline} denote the best and second-best results among specialized long video MLLMs. ``-'' indicates unavailable results. * indicates the average tokens per frame are dynamically adjusted. For our model, we report the theoretical dynamic range (0.5--16) alongside the actual empirical average tokens per frame ({\setlength{\fboxsep}{1pt}\colorbox{gray!10}{\textbf{gray rows}}}), demonstrating that Tempo inherently operates substantially below the maximum budget limits in practice.}
    \label{tab:main_table}
    \resizebox{\linewidth}{!}{
    \begin{tabular}{l c c ccccc}
        \toprule
        \multirow{2.5}{*}{\textbf{Model}} & \multirow{2.5}{*}{\textbf{Size}} & \textbf{Tokens} & \textbf{LongVideoBench} & \textbf{MLVU} & \multicolumn{2}{c}{\textbf{Video-MME (w/o sub.)}} & \textbf{LVBench} \\
        \cmidrule(lr){6-7}
        & & \textbf{per frame} & (473s) & (651s) & Overall (1010s) & Long (2386s) & (4101s) \\
        \midrule
        
        \addlinespace[0.3ex] 
        \multicolumn{8}{l}{\textit{Proprietary Models}} \\
        \addlinespace[0.3ex] 
        GPT-4o~\cite{hurst2024gpt}             &  - &  - & 66.7 & 64.6 & 71.9 & 65.3 & 30.8 \\
        Gemini 1.5 Pro~\cite{team2024gemini}   &  - &  - & 64.0 & -    & 75.0 & 67.4 & 33.1 \\
        \midrule
        
        \addlinespace[0.3ex]
        \multicolumn{8}{l}{\textit{General Open-Source MLLMs}} \\
        \addlinespace[0.3ex]
        VideoChat2-HD~\cite{li2025videochat}    & 7B & 72   & -    & 47.9 & 45.3 & 39.8  & - \\
        LLaVA-OneVision~\cite{li2024llava}      & 7B & 196  & 56.4 & 64.7 & 58.2 & -     & - \\
        LLaVA-Video~\cite{zhang2024llava}       & 7B & 676  & 58.2 & 70.8 & 63.3 & -     & - \\
        VideoLLaMA3*~\cite{zhang2025videollama} & 7B & $\le$ 91 & 59.8 & 73.0 & 66.2 & 54.9 & 45.3 \\
        InternVL3.5~\cite{wang2025internvl3}    & 8B & 256  & 62.1 & 70.2 & 66.0 & -    & - \\
        Molmo2~\cite{clark2026molmo2}           & 8B & 83   & 67.5 & -    & 69.9 & -    & 52.8 \\
        Qwen2.5-VL~\cite{bai2025qwen3}          & 7B & 1924 & 56.0 & 70.2 & 65.1 & -    & 45.3 \\
        Qwen3-VL*~\cite{bai2025qwen3}           & 2B & $\le$ 640 & -    & 68.3 & 61.9 & -    & 47.4 \\
        Qwen3-VL*~\cite{bai2025qwen3}           & 8B & $\le$ 640 & -    & 78.1 & 71.4 & -    & 58.0 \\
        \midrule
        
        \addlinespace[0.3ex]
        \multicolumn{8}{l}{\textit{Specialized Long Video MLLMs}} \\
        \addlinespace[0.3ex]
        LLaMA-VID~\cite{li2024llama}       & 7B   & 2   & -    & 33.2 & 25.9  & -   & 23.9 (13B) \\
        LongVA~\cite{zhang2024long}        & 7B   & 144 & -    & 56.3 & 52.6 & 46.2 & - \\
        Kangaroo~\cite{liu2024kangaroo}    & 8B   & 256 & 54.8 & 61.0 & 56.0 & 46.7 & 39.4 \\
        LongLLaVA~\cite{wang2024longllava} & A13B & 144 & 53.5 & -    & 53.8 & 46.4 & - \\
        LongVILA~\cite{chen2024longvila}   & 7B   & 196 & 57.1 & -    & 60.1 & 47.0 & - \\
        LongVU~\cite{shen2024longvu}       & 7B   & 64  & -    & 65.4 & 60.6 & 59.5 & - \\
        Storm~\cite{jiang2025storm}        & 7B   & 64  & 60.5 & 72.9 & 63.4 & 53.4 & - \\
        BIMBA~\cite{islam2025bimba}        & 7B   & 36 & 59.5 & 71.4 & 64.7 & -    & - \\
        VideoChat-Flash~\cite{li2024videochat}  & 7B  & 16   & \underline{64.7} & 74.7 & 65.3 & 55.4 & 48.2 \\

        \midrule
        \rowcolor{gray!10} \textbf{Tempo* (4K Budget)} & \textbf{6B} & \textbf{0.5--16} & 64.5 & \textbf{75.6} & \textbf{67.8} & \textbf{57.8} & \textbf{52.7} \\
        
        \rowcolor{gray!10} \multicolumn{3}{r}{\textit{\textcolor{darkgray}{$\hookrightarrow$ actual avg. toks/frame}}} & \textit{\textcolor{darkgray}{2.8}} & \textit{\textcolor{darkgray}{2.8}} & \textit{\textcolor{darkgray}{3.6}} & \textit{\textcolor{darkgray}{3.4}} & \textit{\textcolor{darkgray}{2.9}} \\

        \midrule
        \rowcolor{gray!10} \textbf{Tempo* (8K Budget)} & \textbf{6B} & \textbf{0.5--16} & \textbf{65.1} & \underline{75.2} & \underline{67.7} & \underline{57.0} & \underline{52.3} \\
        
        \rowcolor{gray!10} \multicolumn{3}{r}{\textit{\textcolor{darkgray}{$\hookrightarrow$ actual avg. toks/frame}}} & \textit{\textcolor{darkgray}{3.1}} & \textit{\textcolor{darkgray}{3.3}} & \textit{\textcolor{darkgray}{4.3}} & \textit{\textcolor{darkgray}{4.1}} & \textit{\textcolor{darkgray}{3.5}} \\
        \bottomrule
    
    \end{tabular}
    }
\end{table*}
\subsection{Long Video Understanding}
\label{sec:main_results}

Tab.~\ref{tab:main_table} summarizes the evaluation of Tempo (capped at 1024 frames) against state-of-the-art MLLMs across four major benchmarks. Despite a compact 6B-parameter architecture and aggressive token compression (0.5--16 tokens/frame), Tempo achieves state-of-the-art performance. While larger open-weight models (\eg, Qwen3-VL 8B, Molmo2) yield strong absolute scores via exorbitant visual token consumption, Tempo operates under extreme efficiency. By routing evidence through ATA, Tempo strictly bounds visual tokens to 4K or 8K budgets. In practice, ATA dynamically distributes bandwidth so efficiently that the actual consumption falls well below these limits (\eg, 2.9 tokens/frame on LVBench under the 4K budget). Remarkably, its comparative advantage over specialized long video MLLMs amplifies as the temporal span extends.

\paragraph{Dominance in Ultra-Long Video Understanding.}
The most notable results emerge on the extreme-long benchmark LVBench, a rigorous stress test for long-term memory and evidence retrieval. Operating strictly within a 4K visual budget, Tempo achieves \textbf{52.7}, outperforming the strongest 7B specialized MLLM, VideoChat-Flash (48.2), by 4.5 points. Impressively, despite its compact capacity, Tempo eclipses proprietary systems in this ultra-long setting, surpassing GPT-4o (30.8) and Gemini 1.5 Pro (33.1) by massive margins.
This proves that explicit query-aware compression is vastly superior to blindly feeding raw frames into expansive LLM context windows, which often suffer from attention dilution.

\paragraph{Robustness Across Varied Temporal Contexts.}
This dominance consistently extends across other benchmarks. On Video-MME, Tempo secures \textbf{67.8} under the 4K budget, exceeding VideoChat-Flash (65.3) and showing massive improvement over its base model Qwen3-VL-2B (61.9). On the challenging Video-MME \textit{Long} subset (2386s), Tempo achieves \textbf{57.8}. Similarly, Tempo delivers SOTA-level performances on MLVU (\textbf{75.6}) and LongVideoBench (\textbf{65.1} under 8K), asserting its robust generalization across diverse temporal scales and tasks.

\paragraph{{The ``Less is More'' Phenomenon.}} 
Crucially, Tempo's performance under the 4K budget frequently matches or exceeds the 8K budget (\eg, 52.7 vs. 52.3 on LVBench; 57.8 vs. 57.0 on Video-MME \textit{Long} subset). This counter-intuitive phenomenon powerfully validates our ATA strategy. Enforcing a stricter information bottleneck filters out background distractors, forcing the LLM to focus purely on high-value semantic beats. This actively mitigates the \emph{lost-in-the-middle} phenomenon without requiring additional inference compute.

\paragraph{Qualitative Analysis.}
We provide comprehensive qualitative results in Appendix~\ref{supp:qualitative} to further analyze Tempo's adaptive behavior. We contrast localized queries (requiring pinpoint accuracy) with global queries (requiring holistic understanding). These comparisons explicitly demonstrate how Tempo dynamically shifts its compression rhythm---allocating high-fidelity bandwidth to query-critical moments while applying extreme sparsity to irrelevant backgrounds. For overarching queries lacking singular salient events, ATA gracefully defaults to a smooth, low-variance token allocation, ensuring global temporal comprehension remains intact.

\begin{table}[t]
    \centering
    \caption{\textbf{Ablation studies on Tempo's core components.} We decompose our framework across five dimensions: (A) progressive training curriculum, (B) segment-level budget allocation, (C) intra-segment token reduction scheme, (D) relevance scoring source, and (E) temporal continuity. Unless otherwise specified, all variants process videos uniformly sampled at 2 FPS up to a maximum of 1024 frames, strictly bounded by an 8K visual token budget for fair comparison. The default Tempo configuration is highlighted in {\setlength{\fboxsep}{1pt}\colorbox{gray!10}{\textbf{gray}}}. LongVB denotes LongVideoBench.}
    \label{tab:ablation_main}
    \setlength{\tabcolsep}{9pt}
    \resizebox{\linewidth}{!}{
    \begin{tabular}{l c c c c c}
        \toprule
        \multirow{2}{*}{\raisebox{-1.5ex}{\textbf{Ablation Setting}}} & \textbf{LongVB} & \textbf{MLVU} & \multicolumn{2}{c}{\textbf{Video-MME (w/o sub.)}} & \textbf{LVBench} \\
        \cmidrule(lr){4-5}
        & (473s) & (651s) & \begin{tabular}{@{}c@{}}\textbf{Overall} \\ (1010s)\end{tabular} & \begin{tabular}{@{}c@{}}\textbf{Long} \\ (2386s)\end{tabular} & (4101s) \\
        \midrule
        
        \multicolumn{6}{l}{\textit{A. Progressive Training Curriculum}} \\
        w/o Stage 3 - Long-Context SFT (16K Budget)        & 61.4 & 67.2 & 66.1 & 56.3 & 47.3 \\
        w/o Adaptive Token Allocation (16K Budget)         & 62.8 & 73.5 & 67.0 & 56.2 & 51.1 \\
        \rowcolor{gray!10} Tempo Default (8K Budget)       & \textbf{65.1} & \textbf{75.2} & \textbf{67.7} & \textbf{57.0} & \textbf{52.3} \\
        \midrule
        
        \multicolumn{6}{l}{\textit{B. Segment-Level Budget Allocation}} \\
        Uniform Subsampling (Equal tokens per segment)      & 61.9 & 74.0 & 66.3 & 55.2 & 49.9 \\
        Random Drop (Uniform random segment selection)      & 59.3 & 70.9 & 63.6 & 55.2 & 49.8 \\
        Adversarial Routing (Keep lowest-scoring segments)  & 50.7 & 59.3 & 52.4 & 47.8 & 36.9 \\ 
        Hard Top-K Routing (Keep highest-scoring segments)  & 63.5 & 73.9 & 66.7 & 56.2 & 52.7 \\
        \rowcolor{gray!10} ATA (Adaptive token allocation, Alg.~\ref{alg:ata}) & \textbf{65.1} & \textbf{75.2} & \textbf{67.7} & \textbf{57.0} & \textbf{52.3} \\
        \midrule

        \multicolumn{6}{l}{\textit{C. Intra-Segment Token Reduction Scheme}} \\
        Uniform Tail Truncation (Fixed 64 tokens)              & 59.5 & 71.6 & 64.1 & 54.8 & 41.8 \\
        Uniform Head Truncation (Fixed 64 tokens)              & 63.2 & 73.4 & 66.9 & 56.2 & 51.5 \\ 
        Token Merging (Merge visual features to $k_i$ tokens)  & 63.6 & 74.9 & 66.3 & 55.4 & 53.0 \\
        Dynamic Tail Truncation (Keep last $k_i$ tokens)       & 61.9 & 73.4 & 64.8 & 54.2 & 50.5 \\
        \rowcolor{gray!10}Dynamic Head Truncation (Keep first $k_i$ tokens) & \textbf{65.1} & \textbf{75.2} & \textbf{67.7} & \textbf{57.0} & \textbf{52.3} \\
        \midrule
        
        \multicolumn{6}{l}{\textit{D. Relevance Scoring Source \& Zero-Shot Prior}} \\
        Base Model Prior (Standard prompt)                              & 64.6 & 75.1 & 67.2 & 56.1 & 52.6 \\
        Base Model Prior (Explicit routing prompt)                      & 65.7 & 76.3 & 67.6 & 57.3 & 52.7 \\
        External Dense Retriever (Qwen3-VL Reranker) & 64.3 & 75.4 & 67.2 & 57.0 & 51.8 \\
        Tempo SVLM Prior (Standard prompt)                               & 64.1 & 75.4 & 67.2 & 57.0 & 53.4 \\
        \rowcolor{gray!10} Tempo SVLM Prior (Explicit routing prompt)    & \textbf{65.1} & \textbf{75.2} & \textbf{67.7} & \textbf{57.0} & \textbf{52.3} \\
        \midrule

        \multicolumn{6}{l}{\textit{E. Temporal Continuity (Minimum Token Guarantee)}} \\
        Hard Pruning (0 tokens for irrelevant segments)            & 63.9 & 74.8 & 67.4 & 56.3 & 52.3 \\
        \rowcolor{gray!10} Minimal Temporal Anchors ($k_{\min}=4$) & \textbf{65.1} & \textbf{75.2} & \textbf{67.7} & \textbf{57.0} & \textbf{52.3} \\
        
        \bottomrule
    \end{tabular}
    }
    \vspace{-1.0em}
\end{table}

\subsection{Ablation Studies}
\label{sec:exp:ablation}

\begin{figure}[t]
    \centering
    \includegraphics[width=\textwidth]{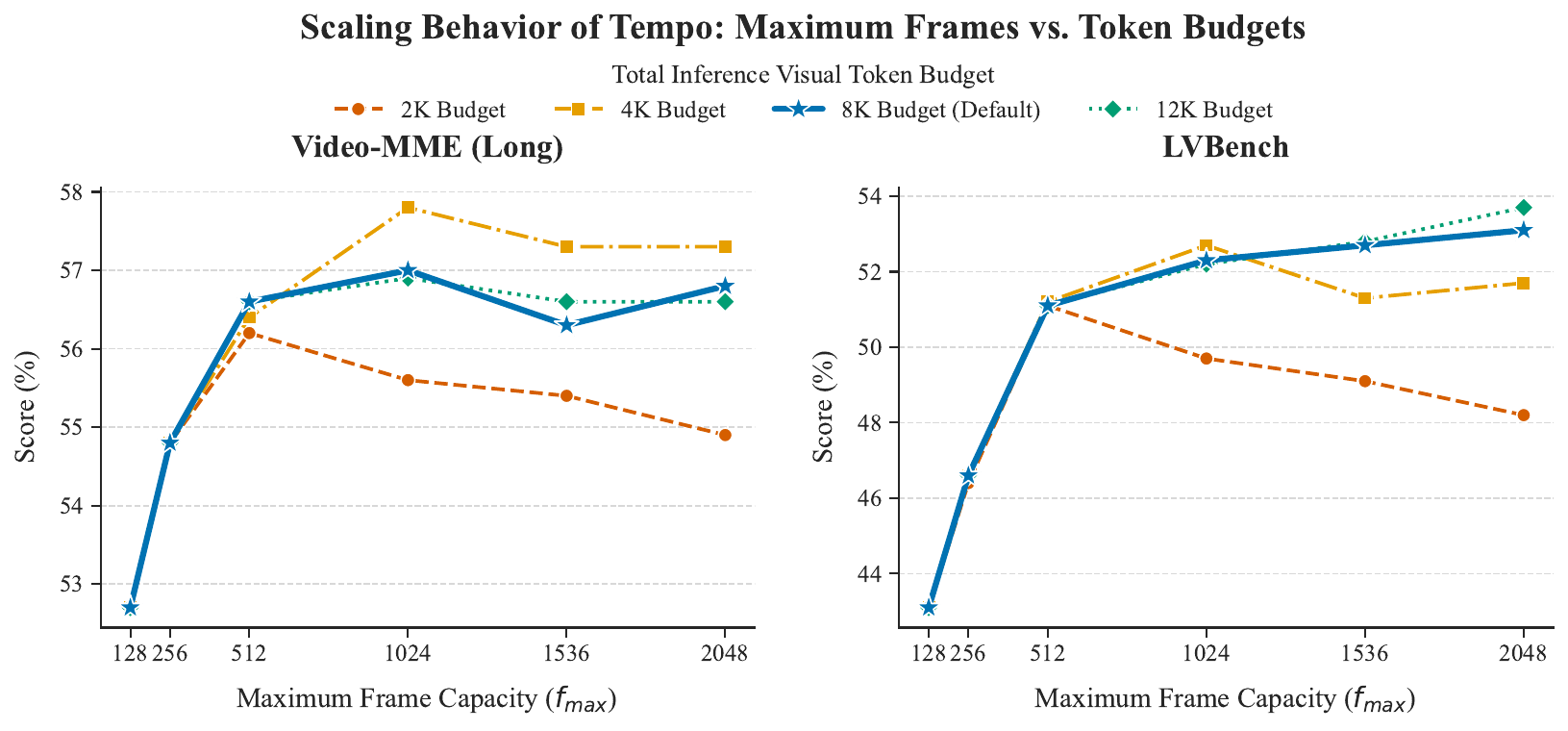}
    \caption{\textbf{Scaling behavior of Tempo.} We investigate the interplay between maximum frame capacity ($f_{\max}$) and total visual token budgets. \textbf{(Left)} On Video-MME (Long), a strict 4K budget acts as an optimal \emph{sweet spot} by aggressively filtering redundancy, whereas larger budgets (8K/12K) yield marginal noise. \textbf{(Right)} On the extreme-long LVBench, restrictive budgets eventually limit the achievable performance, whereas expansive capacities (\eg, 12K) monotonically unlock new peaks at higher frame densities, proving the necessity of scaled context for hour-long video understanding.}
    % \vspace{-1.5em}
    \label{fig:scaling_behavior}
\end{figure}
We systematically decompose Tempo's core components in Tab.~\ref{tab:ablation_main}. Unless specified, all variants process videos uniformly sampled at 2 FPS for a maximum of 1024 frames, strictly bounded by an 8K visual token budget.

\paragraph{A. Progressive Training Curriculum.} 
We first evaluate our training stages (Tab.~\ref{tab:ablation_main}A). Stopping after Stage 2 (w/o Long-Context SFT) yields sub-optimal performance on extreme-long benchmarks (\eg, 47.3 on LVBench), as the LLM fails to extrapolate its temporal window. Introducing Stage 3 without adaptive allocation improves this to 51.1, despite a 16K budget. Crucially, combining the full curriculum with ATA achieves peak performance (52.3 on LVBench) while consuming only half the visual budget (8K). This confirms that extending context is not merely about scaling length, but maximizing information density.

\paragraph{B. Segment-Level Budget Allocation.} 
To strictly satisfy the 8K token budget, all baselines fix the segment capacity at $k_{\max}$ (128) tokens. Given our sampling rate of 2 FPS (capped at a maximum of 1024 frames), Uniform Subsampling halves the initial input frames (processing up to 512 frames). In contrast, Random Drop and the routing variants process all sampled frames (up to 1024) but discard 50\% of the segments to meet the budget (Tab.~\ref{tab:ablation_main}B).
Notably, when we adversarially route by retaining only the lowest-scoring 50\% (Adversarial Routing), performance catastrophically collapses (\eg, 65.1 $\rightarrow$ 50.7 on LongVideoBench). This validates that our zero-shot relevance scores accurately isolate query-critical evidence. Furthermore, while Hard Top-K Routing (dropping the lowest-scoring half entirely) performs competitively, ATA outperforms it by dynamically scaling budgets and preserving overarching causality in most cases.

\paragraph{C. Intra-Segment Token Reduction.} 
We next evaluate the mechanism for token reduction (Tab.~\ref{tab:ablation_main}C). Whether applying a fixed 64-token limit (Uniform) or utilizing our ATA-assigned budgets (Dynamic), Head Truncation consistently eclipses Tail Truncation (\eg, 63.2 vs. 59.5 for Uniform, and 65.1 vs. 61.9 for Dynamic on LongVideoBench). This corroborates our \emph{semantic front-loading} hypothesis (Sec.~\ref{sec:method:ata}): the causal SVLM natively packs the most critical semantics into its earliest memory tokens.
While Token Merging (ToMe) yields a marginal gain on LVBench (53.0 vs. 52.3), it introduces $O(N^2)$ spatial clustering overhead and degrades performance on shorter benchmarks (\eg, 66.3 vs 67.7 on Video-MME). The $O(1)$ dynamic head truncation strikes the optimal balance, offering superior generalizability with strictly zero computational overhead.

\paragraph{D. Relevance Scoring Source \& Zero-Shot Prior.}
We systematically investigate the origin of the relevance scores (Tab.~\ref{tab:ablation_main}D). To isolate the routing effect, all variants here utilize Tempo's compressed memory tokens for final generation, differing only in how the ATA score $s_i$ is computed.
Surprisingly, we observe that the official Qwen3-VL-2B checkpoint (Base Model Prior) possesses a latent yet strong zero-shot capability for relevance alignment, achieving 76.3 on MLVU. However, extracting this prior from the base model, or utilizing an External Dense Retriever (which natively scores segments but yields sub-optimal performance), necessitates a redundant, isolated forward pass per segment.
In contrast, our Tempo SVLM Prior extracts these accurate routing logits simultaneously during the visual compression pass. When guided by an explicit binary instruction (full prompt can be found in the Appendix~\ref{supp:prompts_ablation}), it provides superior comprehensive performance (\eg, 67.7 on Video-MME). This single-pass design capitalizes on the foundational prior with zero latency.

\paragraph{E. Temporal Continuity.}
We ablate the $k_{\min}$ constraint (Tab.~\ref{tab:ablation_main}E). Compared to a Hard Pruning strategy that aggressively drops irrelevant segments to 0 tokens, enforcing Minimal Temporal Anchors (\eg, 4 tokens/segment) consistently improves performance (\eg, 63.9 $\rightarrow$ 65.1 on LongVideoBench). This demonstrates that maintaining a continuous, highly compressed timeline is essential for long-form video understanding, preventing the LLM from losing its temporal orientation and causal tracking in hour-long videos.

\paragraph{Scaling Behavior of Inference Context.}
Fig.~\ref{fig:scaling_behavior} details Tempo's scaling dynamics. At low-frame regimes, all budget curves coincide. As the uncompressed token counts remain below the budget thresholds, ATA allocations are identical. As $f_{\max}$ expands, restrictive budgets (2K, 4K) naturally degrade: enforcing the $k_{\min}=4$ temporal anchor across too many segments prematurely exhausts the capacity, starving query-critical segments of representational bandwidth.
At high frame counts, behavior diverges by video length. Standard long-form tasks (\eg, Video-MME Long) peak at $f_{\max}=1024$ under a 4K budget. Conversely, the extreme-long LVBench necessitates scaled capacities, where expansive 8K and 12K budgets allow performance to scale monotonically, reaching \textbf{53.7} at $f_{\max}=2048$ ($B_{\max}=12$K).
Notably, our empirical profiling (we provide detailed token distributions in the Appendix~\ref{supp:ata_stats}) reveals that even under generous 8K or 12K ceilings, ATA's dynamic sparsity often compresses LVBench videos ($f_{\max}=1024$) to token counts substantially below the allocated budget in practice. This demonstrates a profound property: Tempo allocates bandwidth driven by semantic necessity, rather than greedily padding to fill the available context window.
\section{Conclusion}
\label{sec:conclusion}
We present Tempo, an efficient 6B-parameter framework that resolves the structural mismatch between massive video streams and bounded LLM context windows. Departing from query-agnostic heuristics like sparse sampling or spatiotemporal pooling, Tempo natively unifies a local SVLM and a global LLM. It casts visual token reduction as an early cross-modal distillation process, generating highly compressed, intent-aligned video representations in a single forward pass.
To enforce strict visual budgets at inference without sacrificing fine-grained evidence or overarching causality, we propose Adaptive Token Allocation (ATA). Driven by the SVLM's \emph{zero-shot relevance prior} and empirical \emph{semantic front-loading}, ATA executes $O(1)$ dynamic head truncation. It aggressively routes dense bandwidth to query-critical semantic beats while compressing redundancies into minimal temporal anchors to maintain the global storyline.
Tempo establishes new state-of-the-art performance across diverse benchmarks, notably outperforming specialized long video MLLMs and proprietary baselines on the extreme-long LVBench. Crucially, our scaling analysis reveals that optimal resource allocation depends on video duration. While a compact 4K budget acts as a highly efficient denoiser for standard long video tasks, mastering hour-long narratives necessitates scaled contextual capacities.
By compressing videos to token counts substantially below theoretical limits in practice, Tempo demonstrates a profound property: true long-form multimodal understanding is best achieved not by greedily padding expansive context windows, but through intent-driven, dynamic allocation based on semantic necessity.

\section{Discussion and Future Works}
\label{sec:Discussion}

While Tempo successfully demonstrates that high-efficiency, end-to-end multimodal compression can resolve the context bottlenecks of hour-long videos, this query-aware paradigm opens several critical frontiers for the community. We discuss the current limitations of our framework and possible future research directions.

\paragraph{Eliciting Inherent Relevance Priors via Post-Training.} 
Currently, the Adaptive Token Allocation (ATA) mechanism leverages the SVLM's zero-shot capability to accurately assess whether a video segment is useful for answering a given query. While our ablations demonstrate that this latent prior is already highly effective, it operates entirely in a zero-shot manner. A promising future direction is to explicitly elicit and amplify this inherent capability through post-training (standard supervised fine-tuning risks introducing inductive biases or overfitting to heuristic labels, while reinforcement learning can directly optimize the SVLM's routing policy against the final downstream generation accuracy). By formally optimizing the local compressor to sharpen its internal relevance judgments, we can further elevate this routing precision, potentially driving substantial performance gains across the entire framework.

\paragraph{Autoregressive, Reasoning-Driven Compression.} 
To ensure a highly efficient, single forward pass, Tempo currently compresses video segments using a fixed number of learnable memory tokens. Inspired by recent advancements in reasoning models that dynamically allocate test-time compute (\eg, generating internal thought tokens before halting), a more intelligent paradigm would allow the SVLM to autoregressively generate compressed tokens for a video segment, autonomously deciding when sufficient visual evidence has been gathered to stop. However, adapting this autoregressive extraction without severely bottlenecking inference latency remains a profound optimization challenge for future long video compressors.

\paragraph{Hierarchical On-Demand Distillation for Multi-Turn Dialogue.} 
While Tempo efficiently compresses videos via query-aware distillation, adapting to shifting user intents in multi-turn dialogues currently requires re-extracting visual features from the entire video. A promising frontier is transitioning towards a hierarchical, on-demand routing paradigm. By decoupling a persistent, query-agnostic global context from the intensive query-aware extraction, the global LLM could be empowered to act as an active routing agent. Rather than passively receiving features, the LLM could dynamically identify which specific temporal segments require deeper inspection, invoking the SVLM to distill high-fidelity anchors exclusively for those targeted moments.

\section*{Acknowledgements}
This work was conducted during a research internship at Meta. Junjie Fei, Mingchen Zhuge, Shuming Liu, and Mohamed Elhoseiny were supported by funding from the King Abdullah University of Science and Technology (KAUST) Center of Excellence for Generative AI.

% ---------------------------------------------------------------

% \section{References and citations}

% Take a look at~\cref{table:demo}, appearing on~\cref{section:intro}.
% %
% Some citation of previous work~\citep{goodman}.

\clearpage
\newpage
\bibliographystyle{assets/plainnat}
\bibliography{paper}

\clearpage
\newpage
\beginappendix

\setcounter{figure}{0}
\setcounter{table}{0}
\renewcommand{\thefigure}{\Alph{figure}}
\renewcommand{\thetable}{\Alph{table}}

\section{Statistical Analysis of Adaptive Token Allocation}
\label{supp:ata_stats}

In this section, we present a statistical analysis of the Adaptive Token Allocation (ATA) mechanism to further assess its efficiency and adaptability under diverse video lengths and token budget constraints. The analysis is based on the results reported in Tab.~\ref{tab:main_table}, with all evaluations conducted using \texttt{lmms-eval}. 
For all benchmarks, the temporal context is partitioned into segments with a fixed window size of 8 frames during inference.

\paragraph{Unit Conversion.} Unless otherwise specified, the statistics reported below are measured in \emph{tokens per segment}. The corresponding \emph{average tokens per frame} can be obtained by dividing these values by the window size (8). For example, an allocation of 64 tokens per segment corresponds to 8 tokens per frame.

\subsection{Distribution of Token Allocation}

\begin{figure}[t]
    \centering
    % 4K
    \begin{subfigure}{\textwidth}
        \centering
        \includegraphics[width=0.95\linewidth]{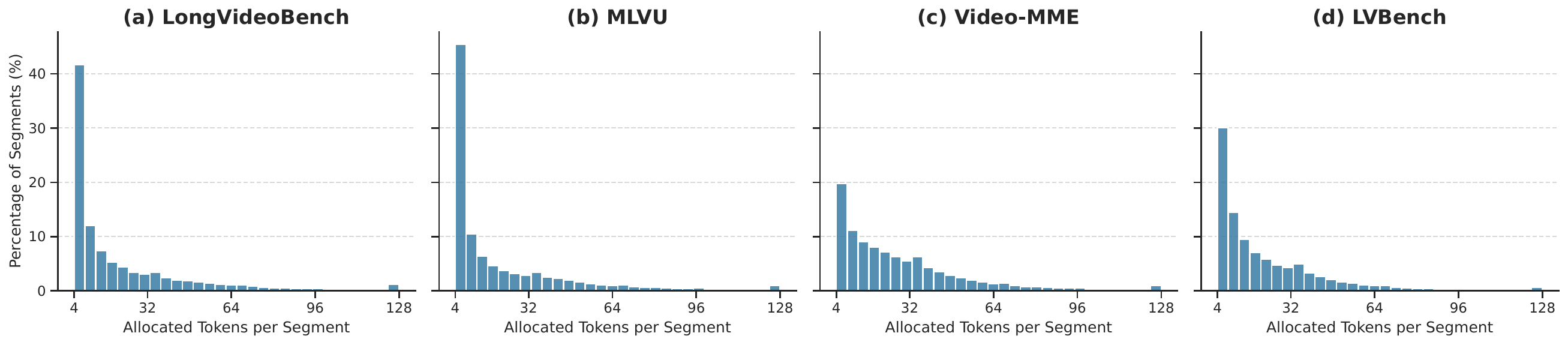}
        \vspace{-0.5em}
    \end{subfigure}
    \\ \vspace{1em}
    % 8K
    \begin{subfigure}{\textwidth}
        \centering
        \includegraphics[width=0.95\linewidth]{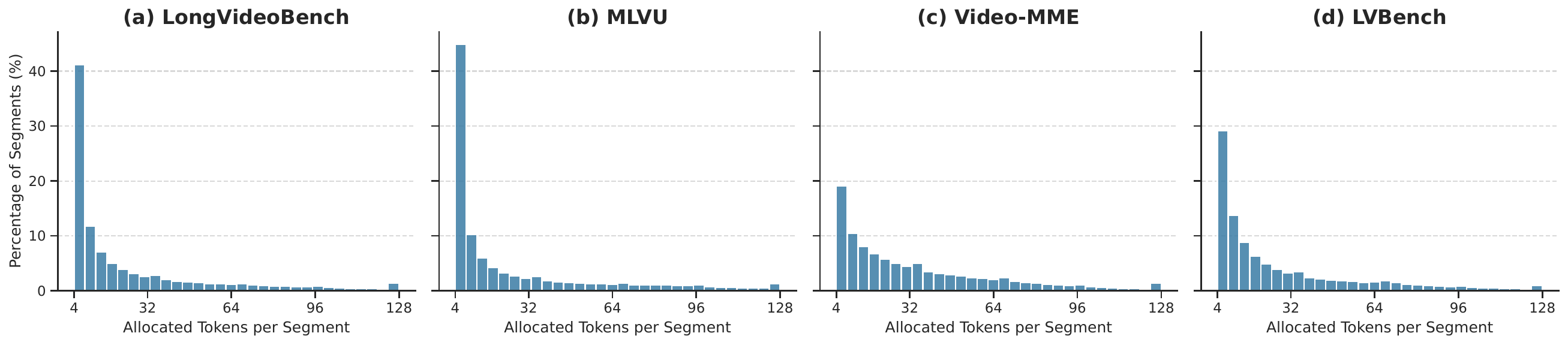}
        \vspace{-0.5em}
    \end{subfigure}
    \caption{
        \textbf{Distribution of allocated tokens per segment.}
        \textbf{(Top)} 4K budget ($B=4096$); \textbf{(Bottom)} 8K budget ($B=8192$).
        Across four benchmarks, ATA consistently exhibits a strongly right-skewed, long-tailed allocation pattern. The majority of segments are compressed into very low-token representations, while a small fraction receives substantially higher allocations for query-aligned segments. Notably, this distribution pattern remains stable under different global budgets.
    }
    \label{fig:micro_dist}
\end{figure}

As shown in Fig.~\ref{fig:micro_dist}, we analyze the behavior of ATA by plotting the percentage of segments against their allocated tokens. The results reveal two key insights:

\begin{itemize}
    \item \textbf{Heavy-Tailed Sparsity:} 
    Across both 4K (Top) and 8K (Bottom) budgets, the allocation exhibits a strongly right-skewed, long-tailed distribution. The most frequent allocations consistently fall into the lowest token bin (corresponding to highly compressed segments), accounting for a large fraction of all segments. This indicates that Tempo compresses redundant content into minimal representations. Meanwhile, the distribution gradually decays toward larger allocations, with a small but consistent fraction of segments approaching the maximum capacity boundary. This suggests that high-fidelity bandwidth is selectively reserved for rare yet query-related segments.
    \item \textbf{Budget Robustness:} 
    The stability of this distribution across different global budgets demonstrates that our SVLM-based compressor produces a consistent, query-driven ranking of visual importance. Rather than uniformly scaling allocations when the budget increases (\eg, from 4K to 8K), Tempo preserves extreme sparsity for background contexts while selectively allocating additional tokens to segments that align with the user's intent.
\end{itemize}

\subsection{Dynamic Budget Utilization and Compression Efficiency}

\begin{figure}[t]
    \centering
    % 4K
    \begin{subfigure}{\textwidth}
        \centering
        \includegraphics[width=0.95\linewidth]{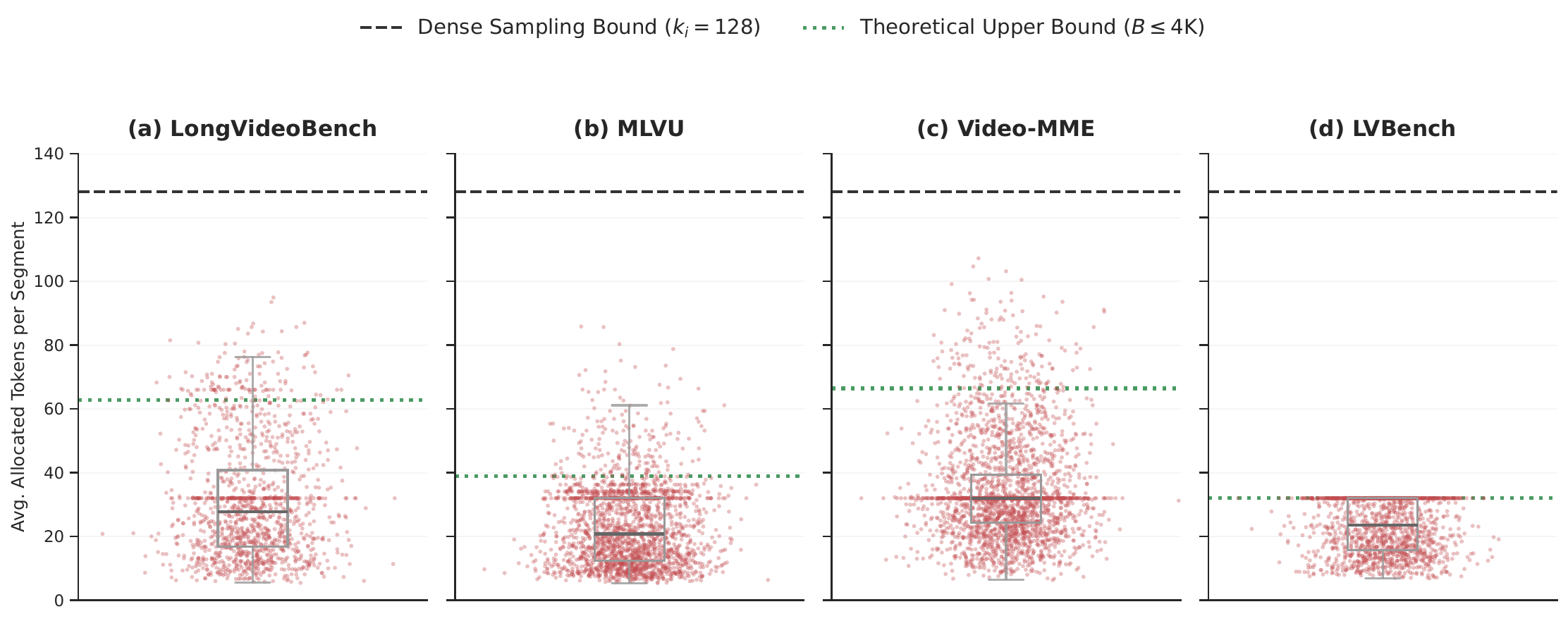}
        \vspace{-0.5em}
    \end{subfigure}
    \\ \vspace{1em}
    % 8K
    \begin{subfigure}{\textwidth}
        \centering
        \includegraphics[width=0.95\linewidth]{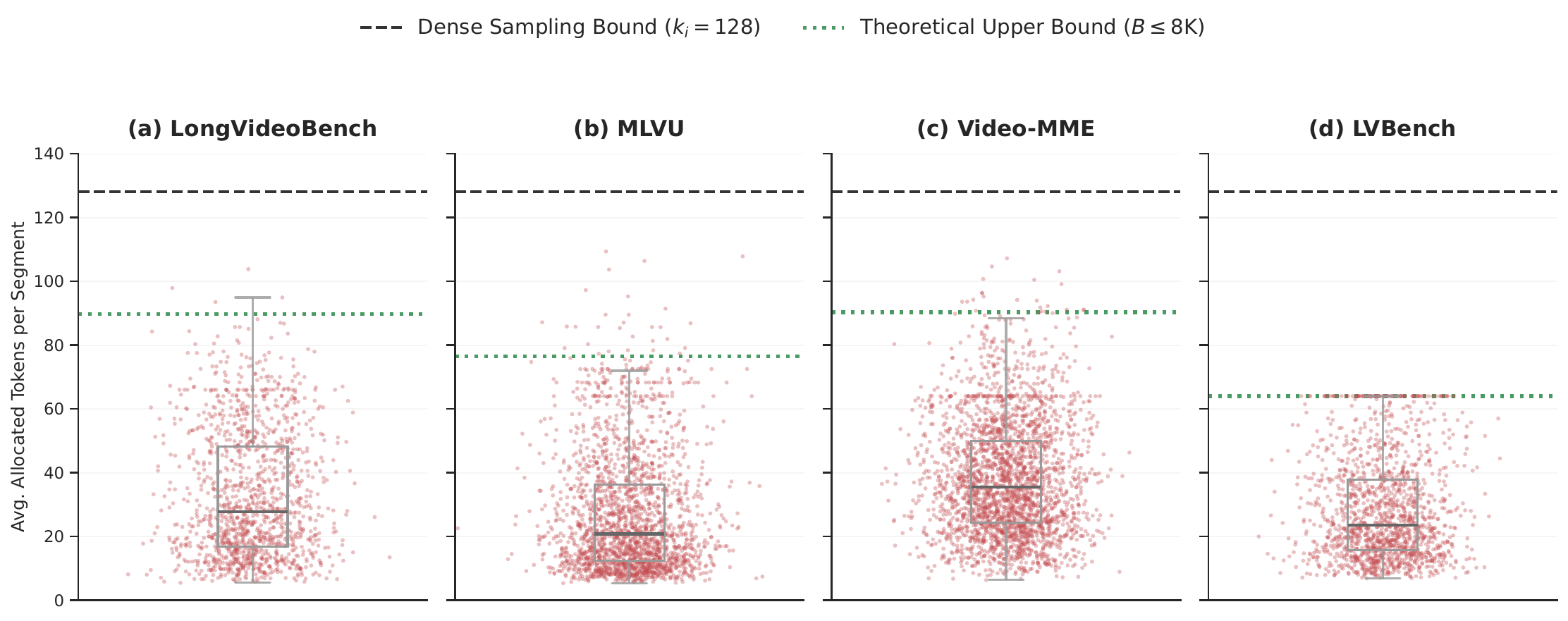}
        \vspace{-0.5em}
    \end{subfigure}
    \caption{
        \textbf{Macro-level budget utilization and adaptation efficiency.}
        \textbf{(Top)} 4K budget; \textbf{(Bottom)} 8K budget.
        Each red dot denotes the \emph{average token consumption per segment} for a video sample. The dashed green line indicates the \emph{dataset-level average theoretical capacity}. Points above the line correspond to shorter videos whose per-segment capacity is higher than the dataset-wide average.
        \textbf{Adaptability:} On datasets with diverse video lengths (\eg, LongVideoBench, Video-MME), the consumption distribution remains well below the theoretical capacity, indicating query-driven compression.
        \textbf{Reliability:} On extremely long videos (LVBench), the token consumption forms a clear ceiling at the theoretical limit, demonstrating strict adherence to the global budget constraint.
    }
    \label{fig:macro_efficiency}
\end{figure}

Fig.~\ref{fig:macro_efficiency} visualizes the macro-level efficiency by comparing the actual average token consumption per segment with the dataset-level average theoretical capacity. 
The green dashed line denotes the \emph{average expected capacity per segment} across the entire dataset, computed as $(N_{\text{samples}} \times B) / \sum S$. Because shorter videos contain fewer segments, their individual theoretical limits are naturally higher than this dataset-wide average. As a result, their actual consumption may exceed the green line without violating the global video budget $B$.

\begin{itemize}
    \item \textbf{Query-Driven Adaptability:} 
    On datasets with diverse video lengths (\eg, LongVideoBench, MLVU, Video-MME), although some shorter videos may peak above the average line, the statistical quartiles (grey spines) and the dense clusters of actual consumption consistently lie well below the dataset-level capacity. This indicates that ATA does not exhaust the available context budget merely because it is available; instead, it preserves bandwidth when the video content is irrelevant to the user query.

    \item \textbf{Hard-Boundary Reliability:} 
    Under extreme long context pressure (\eg, LVBench), where most videos are extremely long and strongly constrained by the maximum segment count, the per-video limits closely align with the dataset-level average. In this regime, the actual consumption forms a clear ceiling at the theoretical limit, demonstrating that Tempo reliably respects the global capacity constraint.
\end{itemize}

\section{Qualitative Analysis of Query-Aware Allocation}
\label{supp:qualitative}

\begin{figure}[p]
    \centering
    \includegraphics[width=0.88\textwidth]{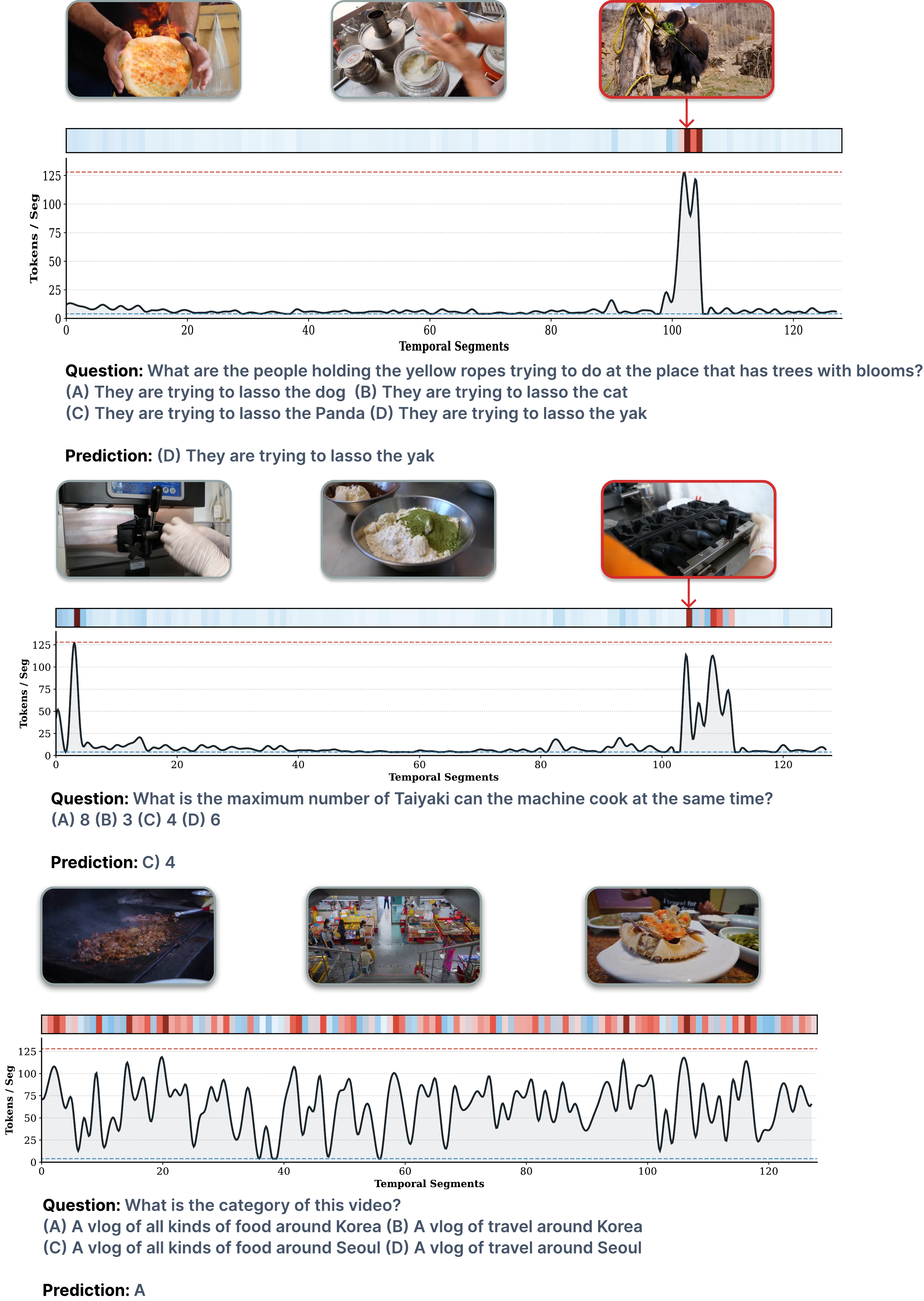}
    \caption{
        \textbf{Qualitative examples of query-aware Adaptive Token Allocation.} We visualize the token allocation distribution across three distinct types of video QA. The red-bordered key frames correspond to segments with high token allocation, while grey-bordered frames represent redundant contexts compressed to minimal tokens. 
        \textbf{(Top) Precise Action Retrieval:} ATA isolates the exact moment the target action (lassoing the yak) occurs. 
        \textbf{(Middle) Targeted Object Grounding:} ATA highlights specific scenes containing cooking machinery relevant to the query concepts (``machine'', ``cook''), while effectively suppressing general manual food preparation. 
        \textbf{(Bottom) Global Summarization:} ATA maintains a dense allocation distribution to capture the overarching theme of the diverse food vlog.
    }
    \label{fig:qual_demo}
\end{figure}

To further illustrate the interpretability and adaptability of the ATA mechanism, we present qualitative visualizations in Fig.~\ref{fig:qual_demo}. These examples show how Tempo dynamically adjusts its temporal context budget according to the semantic requirements of the input query. All examples are sampled from LVBench and illustrate the outputs of \textbf{Tempo 8K}, as reported in Tab.~\ref{tab:main_table}.

\begin{itemize}
\item \textbf{Precise Action Retrieval (Fig.~\ref{fig:qual_demo}, Top):}
For localized tasks requiring specific visual evidence (\eg, identifying what people with yellow ropes are doing), ATA exhibits an extremely sparse allocation pattern. Most irrelevant background segments (\eg, unrelated food preparation) are aggressively compressed into minimal temporal anchors. Conversely, a sharp allocation peak (approaching the maximum budget) is concentrated exactly on the brief segment where the target action (lassoing the yak) occurs.
\item \textbf{Targeted Object Grounding (Fig.~\ref{fig:qual_demo}, Middle):}
When queried about a specific object (\eg, the capacity of a cooking machine), Tempo actively searches for semantically aligned visual cues. High token capacities are dynamically assigned to segments featuring mechanical apparatuses and cooking molds, while manual food preparation scenes are strongly suppressed. Interestingly, even if the localized object visually deviates slightly from a standard Taiyaki maker, this behavior confirms that the allocation is guided by the semantic intersection of the query (``machine'', ``cook'') rather than simple low-level visual saliency.
\item \textbf{Global Video Summarization (Fig.~\ref{fig:qual_demo}, Bottom):}
In contrast, queries requiring holistic understanding (\eg, identifying the overall category of a vlog) demand broader contextual coverage. In this case, ATA avoids extreme sparsification and instead maintains a relatively dense, fluctuating token allocation across the entire temporal sequence, ensuring that distributed thematic cues remain available for reasoning.
\end{itemize}
Overall, these qualitative examples suggest that the local SVLM functions as an effective \emph{smart compressor}. Rather than discarding segments uniformly to satisfy a budget constraint, it performs an interpretable, query-aware cross-modal distillation that prioritizes semantically relevant temporal segments.
\section{Detailed Training Configurations}
\label{supp:training_details}

In the main paper, we introduced a progressive four-stage training curriculum to equip Tempo with long video understanding capabilities. To ensure reproducibility, we provide the detailed hyper-parameter configurations and optimization strategies for all stages extracted directly from our training framework.

\subsection{Optimization and Hardware Setup}

All models are trained on a high-performance compute cluster. Stage 0 (modality alignment) uses 32 NVIDIA H100 (80GB) GPUs, while Stages 1--3 scale to 64 GPUs to accommodate the increased spatiotemporal processing requirements. To support the long sequence lengths of videos and our 6B-parameter architecture, we employ Fully Sharded Data Parallel (FSDP) together with gradient checkpointing. All training is performed in mixed precision (\texttt{bfloat16}) to reduce memory consumption while maintaining numerical stability.
Across all stages, we use the AdamW optimizer with a cosine learning rate scheduler and a 3\% warmup ratio. Weight decay is set to 0.0.

\subsection{Stage-wise Hyper-parameters}

As training progresses from spatial modality alignment (Stage 0) to long-context temporal reasoning (Stage 3), the temporal context length and maximum number of frames are gradually increased. To stabilize this scaling process, different subsets of model parameters are selectively frozen or unfrozen during training. The hyper-parameter configurations for each stage are summarized in Tab.~\ref{tab:hyperparameters}.

\begin{table}[h]
    \centering
    \caption{\textbf{Hyper-parameter configurations across the four-stage training curriculum of Tempo.} \ding{51} indicates that the module is trainable, while \textcolor{gray}{\ding{55}} indicates that it is frozen during training.}
    \label{tab:hyperparameters}
    \resizebox{\textwidth}{!}{
    \begin{tabular}{lcccc}
        \toprule
        \textbf{Configuration} & \textbf{Stage 0: Alignment} & \textbf{Stage 1: Pre-training} & \textbf{Stage 2: Broad SFT} & \textbf{Stage 3: Long-Context SFT} \\
        \midrule
        \multicolumn{5}{l}{\textit{Trainable Modules}} \\
        Local Compressor (SVLM) & \textcolor{gray}{\ding{55} Frozen} & \ding{51} Unfrozen & \ding{51} Unfrozen & \textcolor{gray}{\ding{55} Frozen} \\
        Memory Tokens & \ding{51} Unfrozen & \ding{51} Unfrozen & \ding{51} Unfrozen & \textcolor{gray}{\ding{55} Frozen} \\
        Linear Projector & \ding{51} Unfrozen & \ding{51} Unfrozen & \ding{51} Unfrozen & \textcolor{gray}{\ding{55} Frozen} \\
        Global Decoder (LLM) & \textcolor{gray}{\ding{55} Frozen} & \ding{51} Unfrozen & \ding{51} Unfrozen & \ding{51} Unfrozen \\
        \midrule
        \multicolumn{5}{l}{\textit{Optimization Hyper-parameters}} \\
        LR (LLM \& Projector \& Memory Token) & $1 \times 10^{-3}$ & $1 \times 10^{-5}$ & $1 \times 10^{-5}$ & $1 \times 10^{-5}$ \\
        LR (SVLM) & - & $2 \times 10^{-6}$ & $2 \times 10^{-6}$ & - \\
        Learning Rate Schedule & Cosine & Cosine & Cosine & Cosine \\
        Warmup Ratio & 3\% & 3\% & 3\% & 3\% \\
        Global Batch Size & 128 & 256 & 128 & 64 \\
        Epochs & 1 & 1 & 1 & 1 \\
        \midrule
        \multicolumn{5}{l}{\textit{Data \& Context Scaling}} \\
        Spatial Resolution (Image) & Native & Native & Native & Native \\
        Spatial Resolution (Video) & - & Max 512 (Long edge) & Max 512 (Long edge) & Max 512 (Long edge) \\
        Sampling Rate (FPS) & - & 2 & 2 & 2 \\
        Temporal Window Size & - & 4 frames & 4 frames & 4 frames \\
        Max Sampled Frames ($f_{\max}$) & 1 (Image only) & 8 & 128 & 384 \\
        Segment Capacity ($k_{\max}$) & 128 & 128 & 128 & 128 \\
        LLM Max Context Length & 8192 & 8192 & 8192 & 16384 \\
        \bottomrule
    \end{tabular}
    }
\end{table}

\paragraph{Dynamic Spatial Resolution.}
To balance spatial fidelity with the substantial memory requirements of long temporal contexts, we adopt a dynamic resolution strategy. Static images retain their native resolution to preserve fine-grained visual details. For video samples, frames are resized such that the longest edge does not exceed 512 pixels. The original aspect ratio is preserved without square padding, preventing the SVLM from allocating tokens to redundant regions.

\paragraph{Strategic Freezing.}
During Stage 3 (long-context SFT), we freeze the local SVLM compressor, the memory tokens, and the linear projector. Since this stage focuses on temporal extrapolation, freezing these components isolates the gradient updates to the global LLM. This allows the model to focus on learning long-range temporal dependencies up to 12K visual tokens while avoiding degradation of the learned cross-modal alignment.
\section{Dataset Statistics}
\label{supp:dataset_details}

Tab.~\ref{tab:dataset_summary} summarizes the approximate number of samples per modality (Image, Video, Text) for each training stage. This staged data mixture allows Tempo to gradually transition from spatial alignment on static images to complex spatiotemporal understanding on videos.

\begin{table}[h]
    \centering
    \caption{\textbf{Statistics of training data mixtures.} Approximate number of samples per modality across the four progressive training stages.}
    \label{tab:dataset_summary}
    \resizebox{1.0\textwidth}{!}{
    \begin{tabular}{lcccc}
        \toprule
        \textbf{Modality} & \textbf{Stage 0: Alignment} & \textbf{Stage 1: Pre-training} & \textbf{Stage 2: Broad SFT} & \textbf{Stage 3: Long-Context SFT} \\
        \midrule
        Image Samples & $\sim$558K & $\sim$2.00M & $\sim$0.93M & \multirow{3}{*}{$\sim$384K (Mixed)} \\
        Video Samples & - & $\sim$1.38M & $\sim$2.25M &  \\
        Text Samples  & - & $\sim$143K & $\sim$71K &  \\
        \midrule
        \textbf{Total Size} & \textbf{$\sim$558K} & \textbf{$\sim$3.52M} & \textbf{$\sim$3.25M} & \textbf{$\sim$384K} \\
        \bottomrule
    \end{tabular}
    }
\end{table}
\section{Prompt Design and Relevance Scoring Ablation}
\label{supp:prompts_ablation}

In this section, we describe the system prompts used during training and inference, and analyze their impact on the ATA routing mechanism.

\subsection{System Prompt Definition}

To guide the local SVLM to act as a query-conditioned visual compressor, we design two system prompts:

\vspace{0.5em}
\noindent\textbf{Standard Prompt ($\mathcal{P}_{std}$):}
\begin{quote}
\textit{``You are a query-conditioned visual compressor. Store in the provided memory tokens the minimal visual information needed to answer the Query. Ignore irrelevant details.''}
\end{quote}

\noindent\textbf{Explicit Routing Prompt ($\mathcal{P}_{route}$):}
\begin{quote}
\textit{``You are a query-conditioned visual compressor. Store in the provided memory tokens the minimal visual information needed to answer the Query. Ignore irrelevant details. Now, before compressing, answer exactly `Yes' or `No': is this segment relevant to the Query?''}
\end{quote}

\subsection{Zero-Shot Relevance Prior}

An interesting property of the Tempo SVLM is its \textbf{zero-shot relevance routing capability}. During training, the SVLM is optimized exclusively with the Standard Prompt ($\mathcal{P}_{std}$) to focus on cross-modal compression, and it is \emph{never} explicitly trained to produce binary relevance labels.
At inference time, we instead apply the Explicit Routing Prompt ($\mathcal{P}_{route}$). The SVLM is able to follow this instruction in a zero-shot manner, accurately producing a relevance decision before generating the compressed memory tokens (Sec.~\ref{sec:exp:ablation}B). This behavior suggests that the SVLM possesses the semantic alignment between the query and the visual segment, which can be readily elicited as a routing prior.

\subsection{Ablation Analysis on Relevance Source (Ablation-D)}

To further validate the design, we conduct an ablation study comparing different relevance scoring sources and prompt formulations.

\begin{itemize}

\item \textbf{Impact of Explicit Routing:}
Across both the Base Model Prior and the Tempo SVLM Prior, replacing the standard prompt with the explicit routing prompt ($\mathcal{P}_{route}$) generally improves performance. Encouraging the model to make an explicit relevance decision before context aggregation acts as a useful constraint that helps filter irrelevant segments.

\item \textbf{Observation on Base Model Prior:}
Interestingly, the \emph{Base Model Prior (Standard Prompt)} uses the default system prompt of the Qwen-VL series (\ie \emph{``You are a helpful assistant.''}). Despite the absence of an explicit routing instruction, the model still achieves strong performance. This observation suggests that Qwen-VL models may implicitly assess the relevance between visual inputs and user queries before generating answers.

\end{itemize}
\section{Calculation of Average Tokens per Frame}
\label{supp:a_token_calc}

For a 30-minute video sampled at 2 FPS, the total number of frames is 3,600. Since the shortest videos in LVBench exceed 30 minutes, the raw frame count of each sample is substantially larger than our maximum sampling limit $f_{\max}$. Therefore, we estimate the theoretical upper bound of the average number of tokens per frame as:
\begin{equation}
    \text{Tokens per Frame}_{\max} = \frac{B_{\max}}{f_{\max}},
\end{equation}
where $f_{\max}$ denotes the maximum number of sampled frames provided to the model, and $B_{\max}$ denotes the global visual token budget (context length).

In Fig.~\ref{fig:teaser}(c), we report Tempo under two configurations: a 4K token budget with $f_{\max}=1024$, and a 12K token budget with $f_{\max}=2048$. Although ATA dynamically compresses the visual sequence—often resulting in substantially lower token usage in practice—the \emph{theoretical upper bound} corresponds to the scenario where the full global budget is consumed.
Because the sampled frame count for LVBench is fixed at $f_{\max}$, the resulting upper bounds are $4096/1024 = 4$ and $12288/2048 = 6$ tokens per frame, respectively.
For the baseline VideoLLaMA3~\cite{zhang2025videollama}, the official configuration uses $f_{\max}=180$ and $B_{\max}=16\text{K}$, yielding a theoretical upper bound of approximately 91 tokens per frame.
For LLaMA-ViD~\cite{li2024llama}, the reported results are taken directly from the official LVBench leaderboard.
For all other baselines shown in Fig.~\ref{fig:teaser}(c), we adopt the metrics reported in VideoChat-Flash~\cite{li2024videochat}.

\end{document}